\documentclass{article}
\usepackage{amsmath,graphicx,mlspconf}
\usepackage{algorithm}
\usepackage{algpseudocode}
\usepackage{amsfonts}
\usepackage{caption}
\usepackage{cite}
\usepackage{amssymb,amsfonts}
\usepackage{subcaption}
\usepackage{textcomp}
\usepackage{tikz}
\usepackage{xcolor}

\DeclareMathOperator*{\argmax}{arg\,max}

\copyrightnotice{979-8-3503-2411-2/23/\$31.00 {\copyright}2023 IEEE}

\toappear{Preprint submitted to IEEE MLSP 2023}

\title{SkelEx and BoundEx: Natural Visualization of ReLU Neural Networks}
\name{Pawel Pukowski, Haiping Lu}
\address{Department of Computer Science, University of Sheffield, Sheffield, UK}

\begin{document}

\maketitle

\begin{abstract}
Despite their limited interpretability, weights and biases are still the most popular encoding of the functions learned by ReLU Neural Networks (ReLU NNs). That is why we introduce SkelEx, an algorithm to extract a skeleton of the membership functions learned by ReLU NNs, making those functions easier to interpret and analyze. To the best of our knowledge, this is the first work that considers linear regions from the perspective of critical points. As a natural follow-up, we also introduce BoundEx, which is the first analytical method known to us to extract the decision boundary from the realization of a ReLU NN. Both of those methods introduce very natural visualization tool for ReLU NNs trained on low-dimensional data.
\end{abstract}

\section{Introduction}

ReLU is a very popular activation function not only because of its great performance, but also because of it's piecewise linear (PL) structure. ReLU being PL implies that the learned membership functions are also PL. This PL structure is most commonly used to study the number of linear regions that an architecture can produce, giving us insight on it's flexibility. However, it can have other uses. For example, it allows us to introduce a different encoding of the same functions. Encoding that uses weights and biases, like Eqn. (\ref{eq:1}) is great for training, and space saving, but lacks a lot in terms of interpretability. If we used the information on the location of gradient changes (critical points), like in Eqn. (\ref{eq:2}), the functions would become easier to interpret. Additionally, this \textit{skeleton encoding} introduces a very natural visualization of ReLU NNs trained on low-dimensional data.
\begin{equation} \label{eq:1}
    f(x) = \max(0,\; \max(0,\; x+1) - 2\max(0,\; x))
\end{equation}

\begin{equation} \label{eq:2}
    f(x) =
    \begin{cases}
        0   & x \in (-\infty, -1]\\
        x   & x \in (-1, 0]\\
        -x  & x \in (0, 1]\\
        0   & x \in (1, \infty) 
    \end{cases}
\end{equation}

\begin{figure*}[t]
    \centering
    \includegraphics[width=17.8cm]{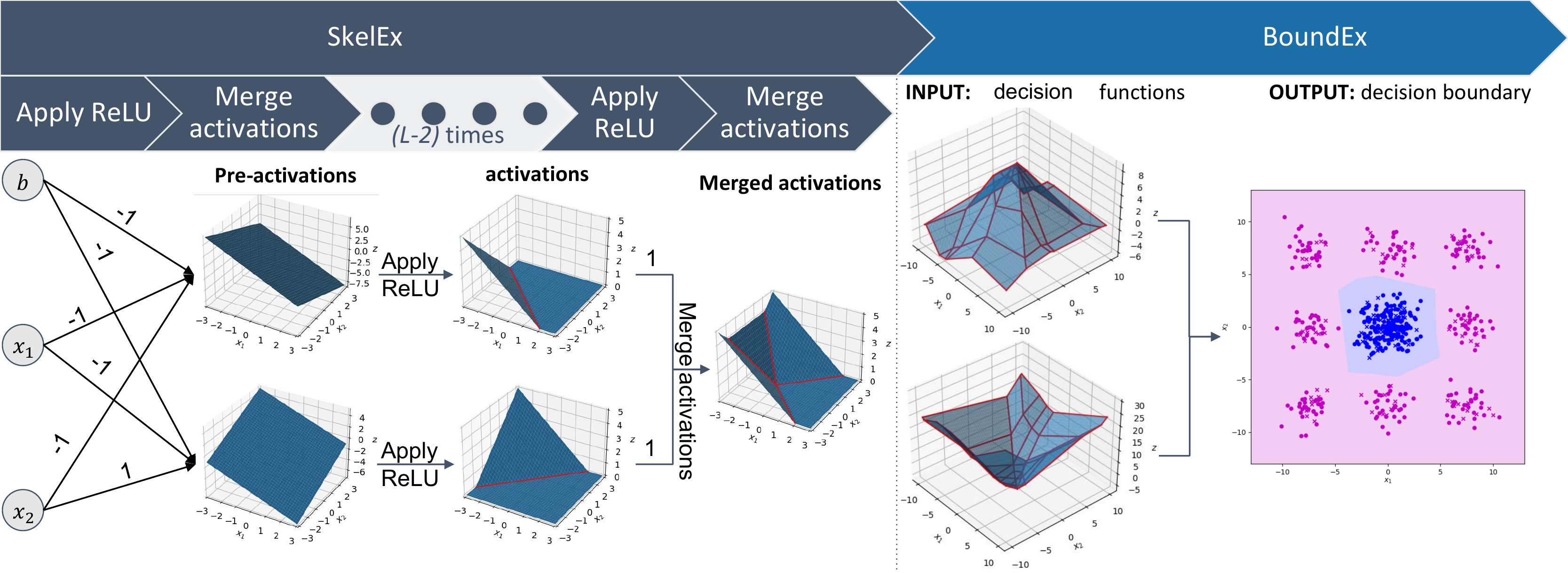}
    \caption{Let $\mathbf{w}=[[[-1, -1], [-1, 1]], [[1, 1]]]$, and $\mathbf{b} = [[-1, -1], [0]]$ be the weights and biases learned by a ReLU NN with configuration 2-2-1 (left). SkelEx($\mathbf{w}, \mathbf{b})$ first extracts pre-activations of the first hidden layer, and then calculates the activations. Next, SkelEx merges the activations to calculate pre-activations of the next layer. For deeper networks this 2 step process would be repeated until the skeletons of the membership functions are extracted (red lines indicate critical points). BoundEx uses the skeletons extracted via SkelEx to calculate the decision boundary (right side; dots indicate training data).}
    \label{fig:poster}
\end{figure*}

Decision boundary is a very important property of the trained NNs, as it is responsible for tessellating the input space into membership regions. Analysing the decision boundary gives us crucial information, such as the exact location of all adversarial examples. However, as far as we know, so far only analytical means of calculating the decision boundary were used. The decision boundary is a subset of the intersection of the membership functions. This means that the skeleton encoding allows us to introduce the first analytical method that extracts the decision boundary of ReLU NNs. Our main contributions are as follows:
\begin{itemize}
    \item We provide an algorithm (SkelEx) that transforms the weights and biases encoding of functions learned by ReLU NNs into the skeleton encoding (Fig. \ref{fig:poster} left).
    \item We visualize the difference between linear regions and activation regions, two entities that have often been incorrectly used as synonyms in literature.
    \item We introduce an analytical method to extract the decision boundary learned by FC ReLU NNs. (Fig. \ref{fig:poster} right)
\end{itemize}

\section{Literature Review}

Linear regions are commonly used to measure the expressivity of ReLU NNs \cite{gamba2022all}. Researchers initially focused on establishing an upper bound on the maximum number of linear regions an architecture could produce \cite{montufar2014number}, and later improving it \cite{raghu2017expressive, serra2018bounding}. The number of linear regions produced is believed to indicate the architecture's flexibility, with more regions leading to smaller approximation errors \cite{liu2021optimal}. However, the average number of linear regions produced by a realization after training is much lower than the upper bound \cite{serra2018bounding}.

Linear regions can be counted numerically or analytically. Numerical methods involve stepping along a trajectory and deciding whether each point belongs to the same linear region as the previous one based on their activations \cite{novak2018sensitivity, raghu2017expressive}. Analytical methods are more elaborate and computationally expensive. Serra et al. \cite{serra2018bounding} counted the number of linear regions across the entire input space, while Hanin et al. \cite{hanin2019deep, hanin2019complexity} introduced properties of linear regions and a method to count them along a 1D or 2D subspace. Zhang et al. \cite{zhang2020empirical} extracted the H-representation of the linear region containing a given input.

Numerical methods are commonly used to obtain the decision boundary for NNs, with the most popular approach being to traverse a defined trajectory away from a data sample in fixed-sized steps until the membership of the point changes \cite{zhang2020empirical, he2018decision, ortiz2020hold, guan2020analysis, karimi2020decision}. Another popular approach is to approximate the decision boundary using adversarial examples \cite{ortiz2020hold, guan2020analysis, karimi2020decision}.

Takahashi et al. \cite{takahashi2011construction} and Zengin et al. \cite{zengin2022super} use convex hulls and Voronoi tessellation, respectively, to classify data samples, but their accuracy is lower than the weights and biases approach because they learn different functions.

\begin{figure}[t]
    \begin{subfigure}{4cm}
        \centering
        \includegraphics[width=4cm]{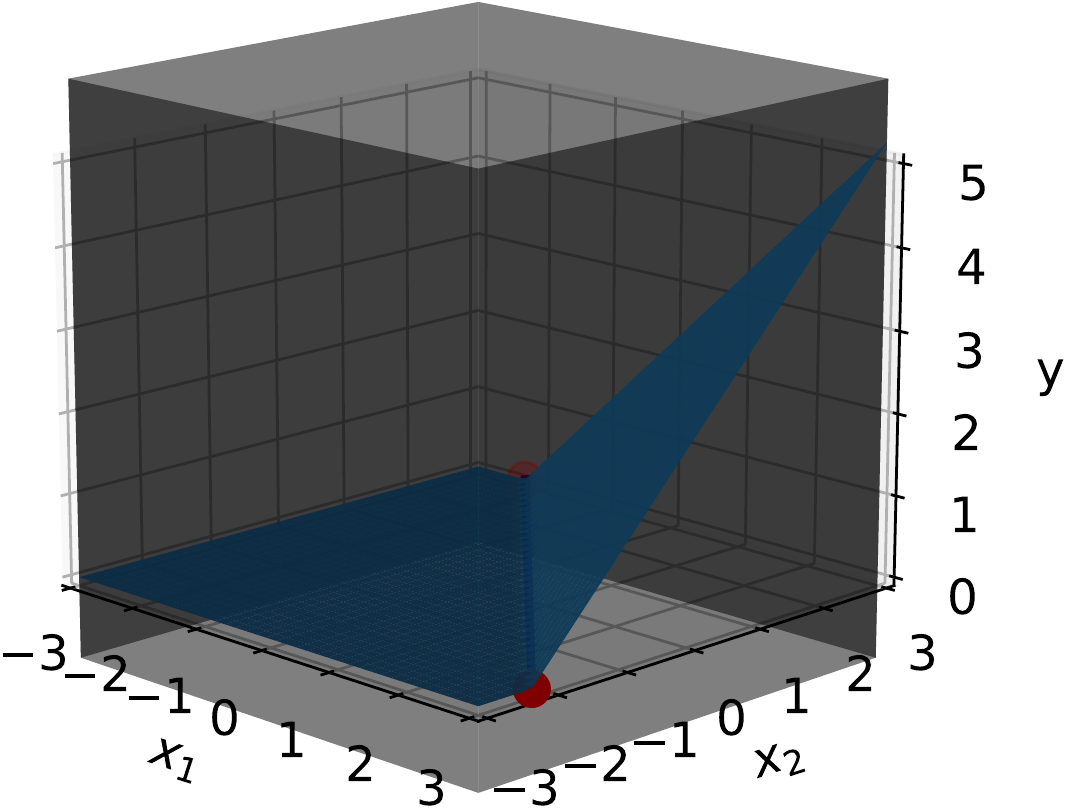}
        \caption{Bounded 3D ReLU}
        \label{fig:bounded_3d_relu}
    \end{subfigure}
    \begin{subfigure}{4cm}
        \centering
        \includegraphics[width=4cm]{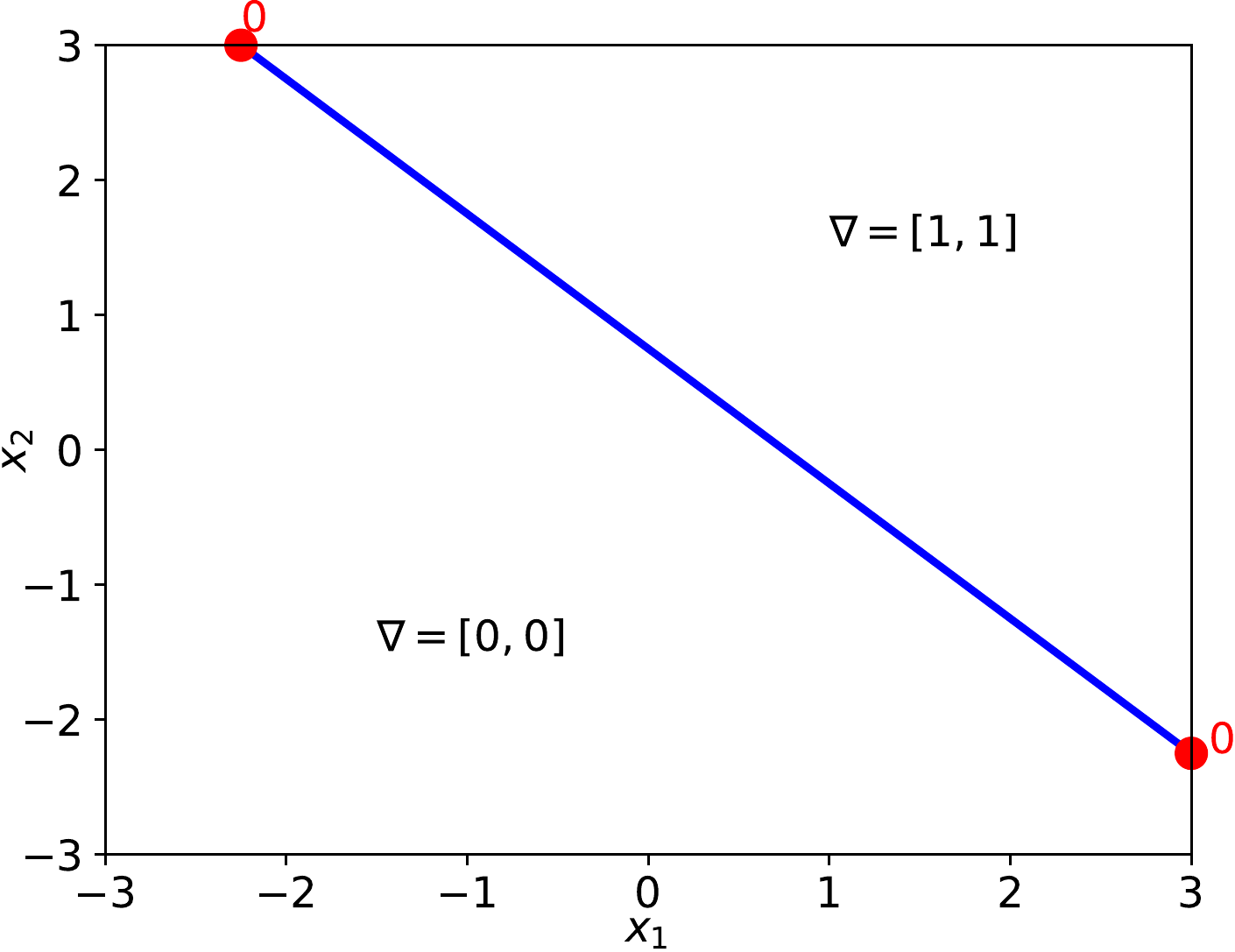}
        \caption{2D representation}
        \label{fig:2d_result}
    \end{subfigure}
    \caption{(a) The hyperrectangle $R$ bounds the input space of $f$. (b) $R$ allows us to encode $f$ using the skeleton of critical points (red points and blue line) and the gradients ($\nabla$) of the linear regions.}
    \vspace{-3mm}
    \label{fig:hyperrectangle_visualization}
\end{figure}

\section{SkelEx}
\label{sec:3}

SkelEx is a two-step algorithm that extracts the skeletons of learned membership functions from pre-trained ReLU NN's weights and biases. It applies ReLU to neurons' pre-activations and merges activations to calculate pre-activations fo the next layer until the final membership functions are obtained (Algorithm \ref{alg:skelex}). SkelEx's output provides a natural visualization of learned functions and information such as critical point formation time (at which layer) and how linear regions tessellate the input space. 

\begin{algorithm}
    \caption{SkelEx}\label{alg:skelex}
    \begin{algorithmic}
        \Require $n_i \gets$ number of neurons in $i^{th}$ layer
        \For{$i \in \{1, 2, ..., L\}$}
            \For{$j \in \{1, 2, ..., n_i\}$}
                \State $f_i^j \gets$ ApplyReLU$(g_i^j)$
            \EndFor
            \For{$j \in \{1, 2, ..., n_{i+1}\}$}
                \State $g_{i+1}^j \gets$ MergeActivations$(\mathbf{f}_i \cdot \mathbf{w}_i)$
            \EndFor
            \State $\mathbf{g}_{i+1} \gets \mathbf{g}_{i+1} + \mathbf{b}_{i+1}$
        \EndFor
        \State \Return $\mathbf{f}_{L+1}$  \Comment{Vector of $k$ membership functions}
    \end{algorithmic}
\end{algorithm}
\vspace{-5mm}

\subsection{$2D$ projection}

The functions learned by FC ReLU NNs reside in $(n_0+1)$-dimensional space, where $n_0$ is the number of input neurons. To encode them we only require the skeletons formed by the membership functions, which reside in $n_0$-dimensional space, and the gradient of linear regions. In Fig. \ref{fig:poster} we used 3D projection, and we can see that even for very easy functions it is already hard to visualize them on 2D paper/screen. That is why from this point on we will visualize all functions using only $n_0$ dimensions. As an example, Fig. \ref{fig:2d_result} shows a $2D$ projection of the function from Fig. \ref{fig:bounded_3d_relu}.

\begin{figure}[t]
    \begin{subfigure}{4cm}
        \centering
        \includegraphics[width=4cm]{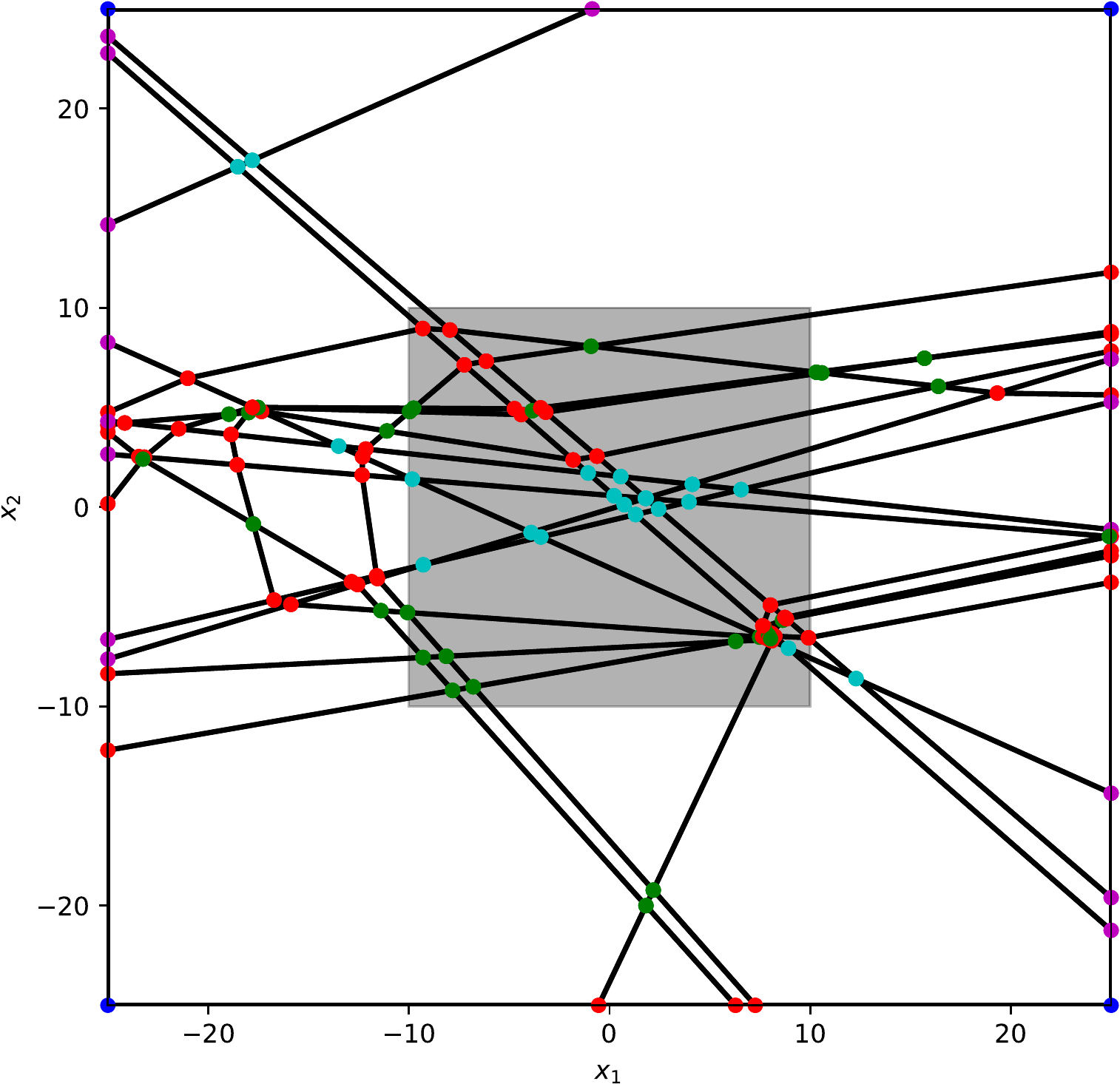}
        \caption{Big hyperrectangle}
        \label{fig:big_hyperrectangle}
    \end{subfigure}
    \begin{subfigure}{4cm}
        \centering
        \includegraphics[width=4cm]{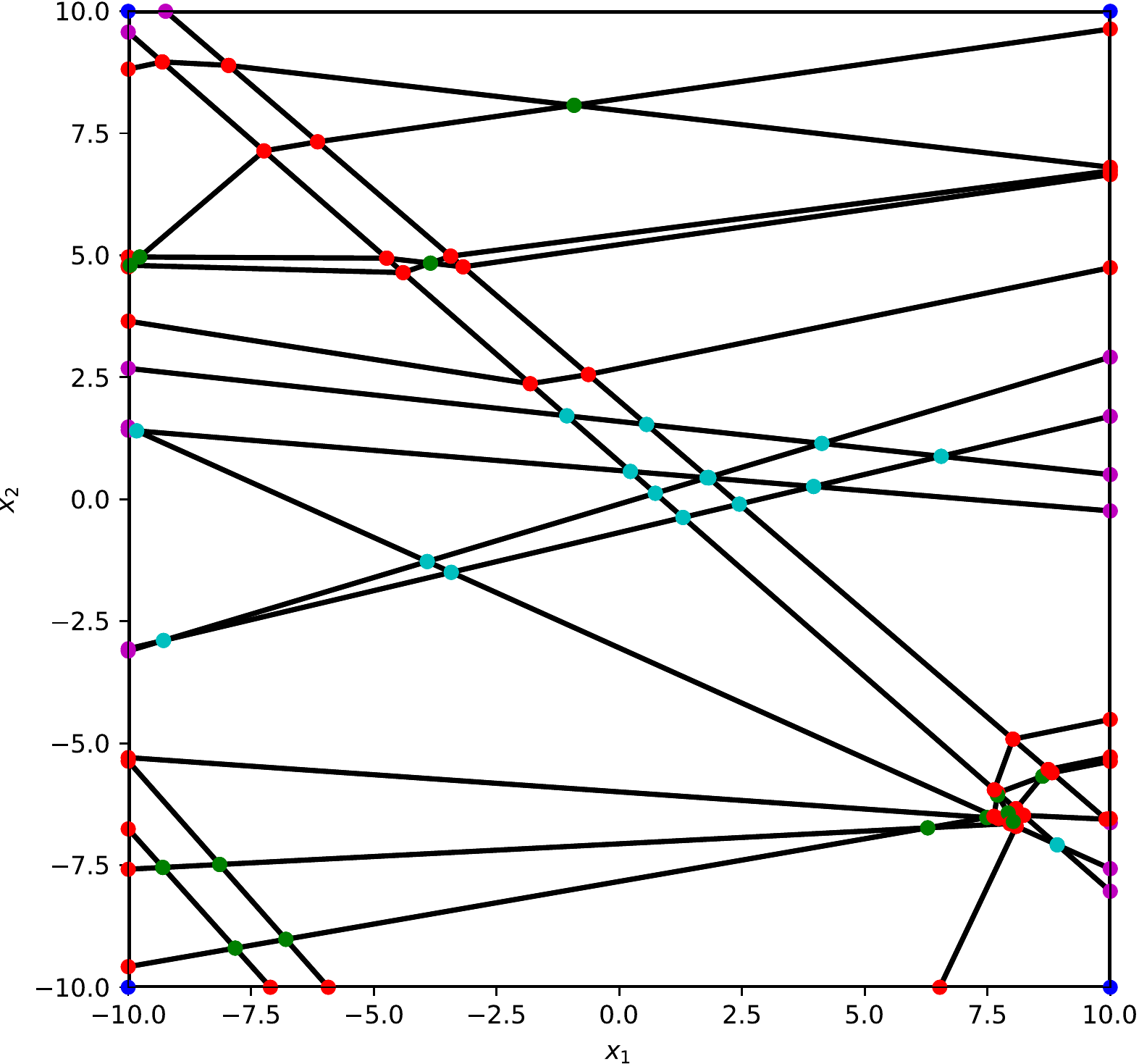}
        \caption{Small hyperrectangle}
        \label{fig:small_hyperrectangle}
    \end{subfigure}
    \caption{SkelEx extracted (a) and (b) using hyperrectangles of different sizes: (a) $x_1, x_2 \in [-25, 25]$ (b) $x_1, x_2 \in [-10, 10]$. (b) is the grey patch from (a). Dots are colored by the step in which they were created (blue - $\mathbf{g}_1$, magenta - $\mathbf{f}_1$, cyan - $\mathbf{g}_2$, red - $\mathbf{f}_2$, green - $\mathbf{g}_3$).}
    \vspace{-3mm}
    \label{fig:hyperrectangle_theorem}
\end{figure}

\subsection{Assumption}

To create a skeleton for the functions learned by ReLU NNs, we must first bind the input space using a hyperrectangle $R$. Despite concerns that this could affect the accuracy of the algorithm, the dimensions of $R$ do not impact the accuracy of SkelEx, as it works with linear regions one at a time. Therefore, restricting the input space simply reduces the number of linear regions or makes the algorithm work ob a subspaces of linear regions. Since most real-world datasets are already bounded, defining the dimensions of the hyperrectangle is usually straightforward (e.g., $age\in [0, 125]$ and $pixel_{MNIST} \in [0, 255]$).

\subsection{Step 1. Applying ReLU}

Let $g_i^j(\mathbf{x})$ be the pre-activation of the $j^{th}$ neuron in the $i^{th}$ layer. We know that, just like the membership function, $g_i^j(\mathbf{x})$ is a PL function. This means that its domain is divided into linear regions. We also know that applying ReLU to a union of linear regions yields the same result as when applying ReLU to each linear region separately and then taking the union. Hence, when calculating ReLU($g_i^j(\mathbf{x})$) we can just go through all linear regions of $g_i^j(\mathbf{x})$, and apply ReLU to each of them, and then merge the results. 

\begin{figure}[t]
    \begin{subfigure}{4cm}
        \centering
        \includegraphics[width=4cm]{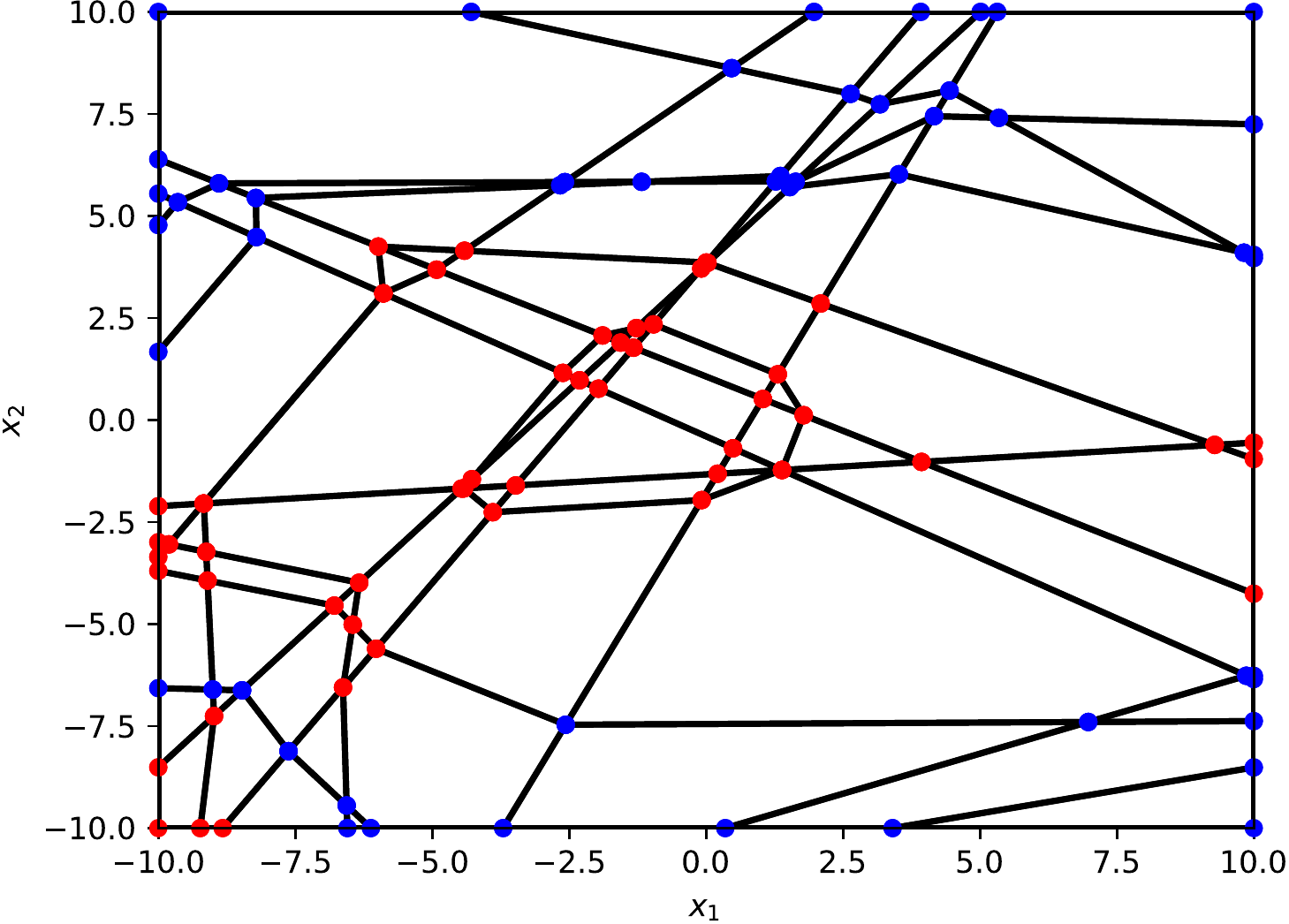}
        \caption{Before ReLU}
        \label{fig:skeleton_before_relu}
    \end{subfigure}
    \begin{subfigure}{4cm}
        \centering
        \includegraphics[width=4cm]{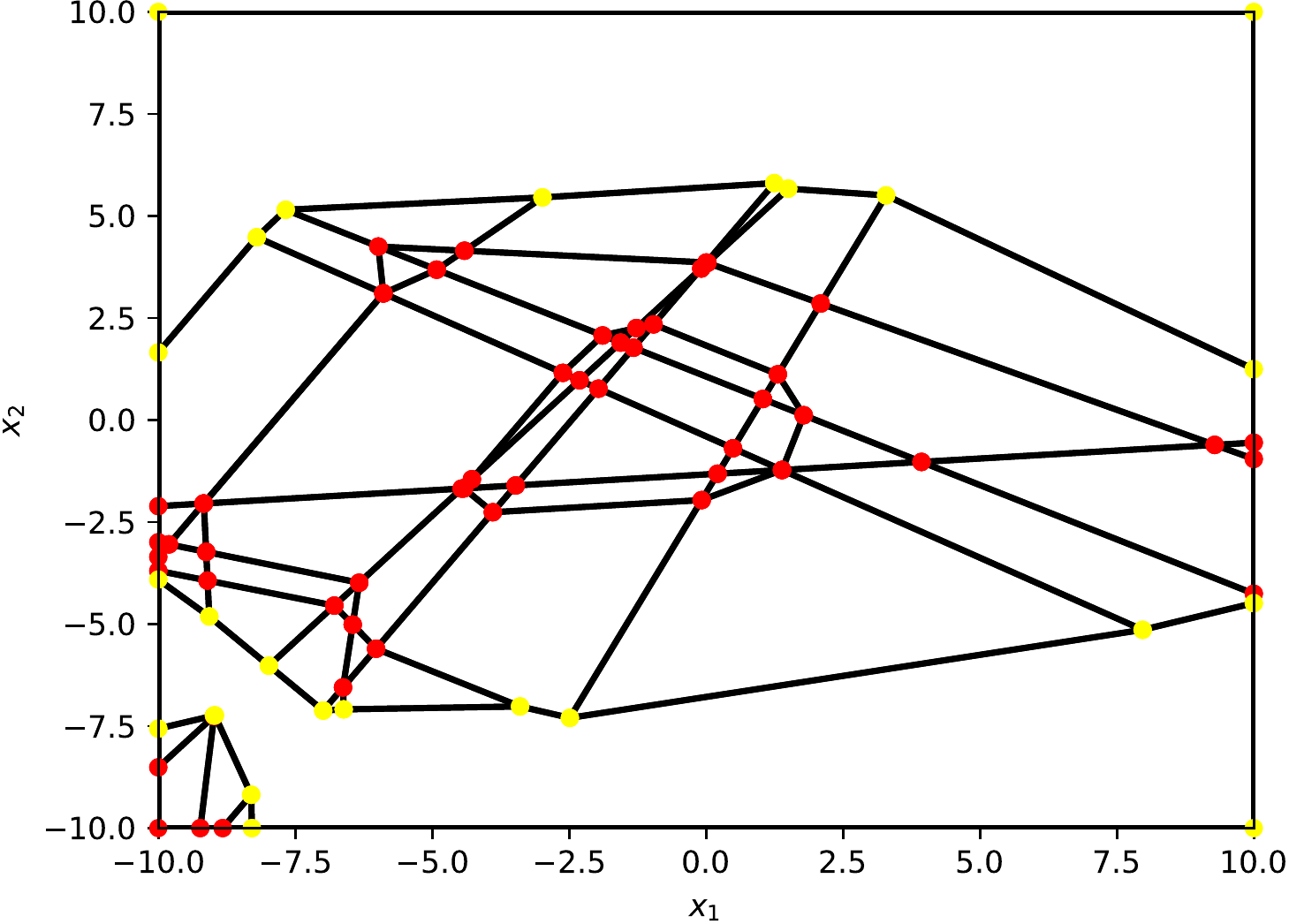}
        \caption{After ReLU}
        \label{fig:skeleton_after_relu}
    \end{subfigure}
    \caption{When ReLU is applied to the skeleton in (a), all negative (blue) vertices are removed, and the positive (red) ones are left intact. At the function's intersection with $0$ ($(x_1, x_2)$-axis) new (yellow) vertices are formed.}
    \label{fig:example_of_applying_relu}
\end{figure}

ReLU splits each linear region into a sequence of regions. We divide those regions into the ones that lie above and below $0$ in the output space. All the regions below $0$ will be moved up to $0$. We will also merge all of the neighboring regions with the same gradient. The whole process is described in Algorithm \ref{alg:apply_relu}, and the visualization is provided in Fig. \ref{fig:example_of_applying_relu}. The merging process is necessary to keep operating on linear regions. If we omitted it, then we would be working on activation regions greatly increasing the computational complexity. 

\begin{algorithm}
    \caption{ApplyReLU}\label{alg:apply_relu}
    \begin{algorithmic}
        \Require $S \gets \{lr_1, ..., lr_n\}$  \Comment{$n$ linear regions tessellate $S$}
        \State $new\_lrs \gets$ empty set
        \For{$lr$ in $S$}  \Comment{Apply ReLU to each $lr$}
            \State $vertices \gets lr.vertices$
            \For{$e \in lr.edges$}
                \If{$sign(e.start) \neq sign(e.end)$}
                    \State $intersection \gets v \subset e$ such that $v.value = 0$
                    \State $vertices.add(intersection)$
                \EndIf
            \EndFor
            \State change values of all negative vertices to $0$
            \State $new\_lrs.add($FormLinearRegions$(vertices))$
        \EndFor
        \State merge lrs in $new\_lrs$ that have the same gradient
        \State \Return $new\_lrs$  \Comment{Convert to Skeleton class}
    \end{algorithmic}
\end{algorithm}

\subsection{Step 2. Merging Activations}

\begin{figure}[t]
    \begin{subfigure}{2.6cm}
        \centering
        \includegraphics[width=2.6cm]{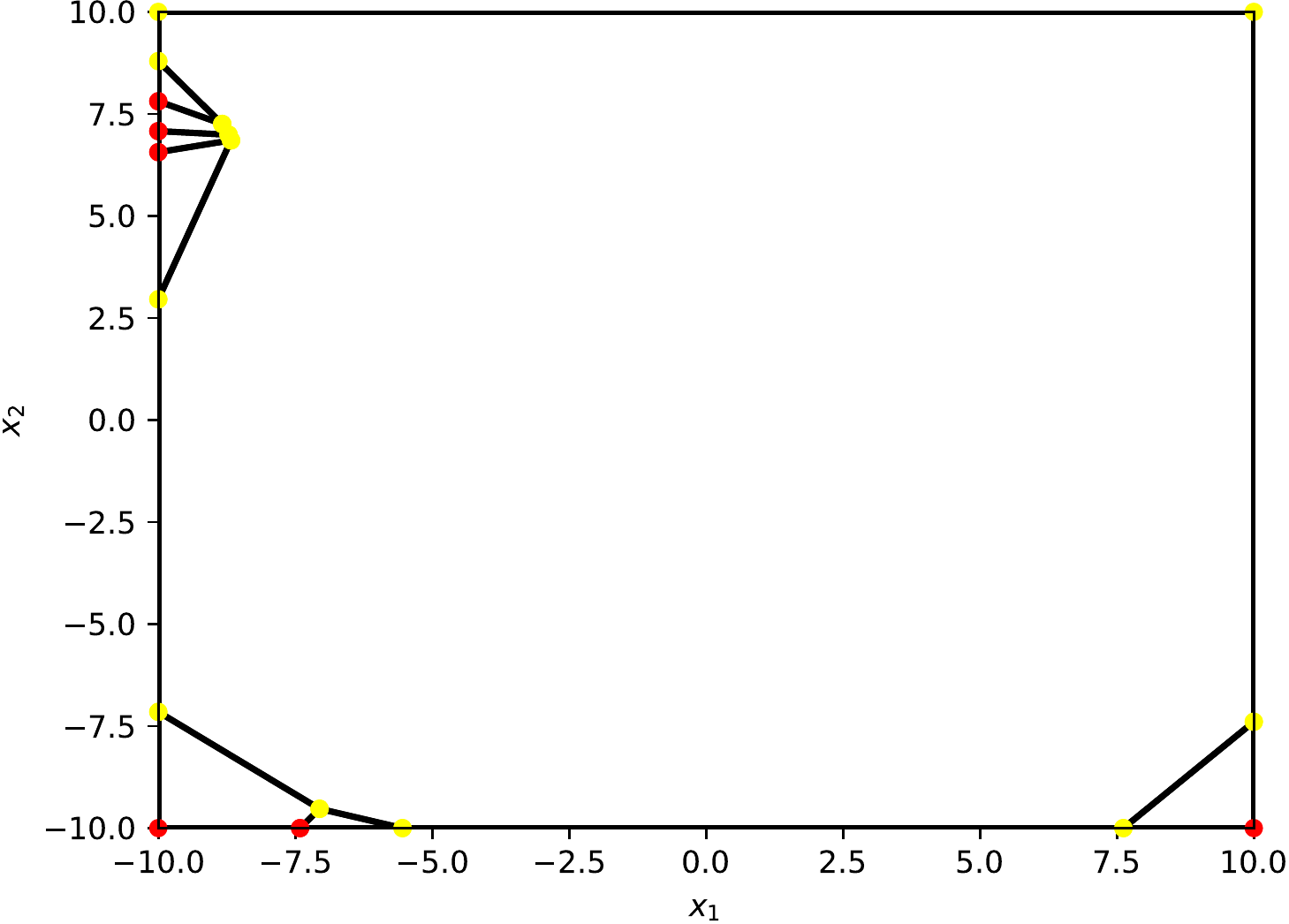}
        \caption{$f^1$}
        \label{fig:skeleton1}
    \end{subfigure}
    \begin{subfigure}{2.6cm}
        \centering
        \includegraphics[width=2.6cm]{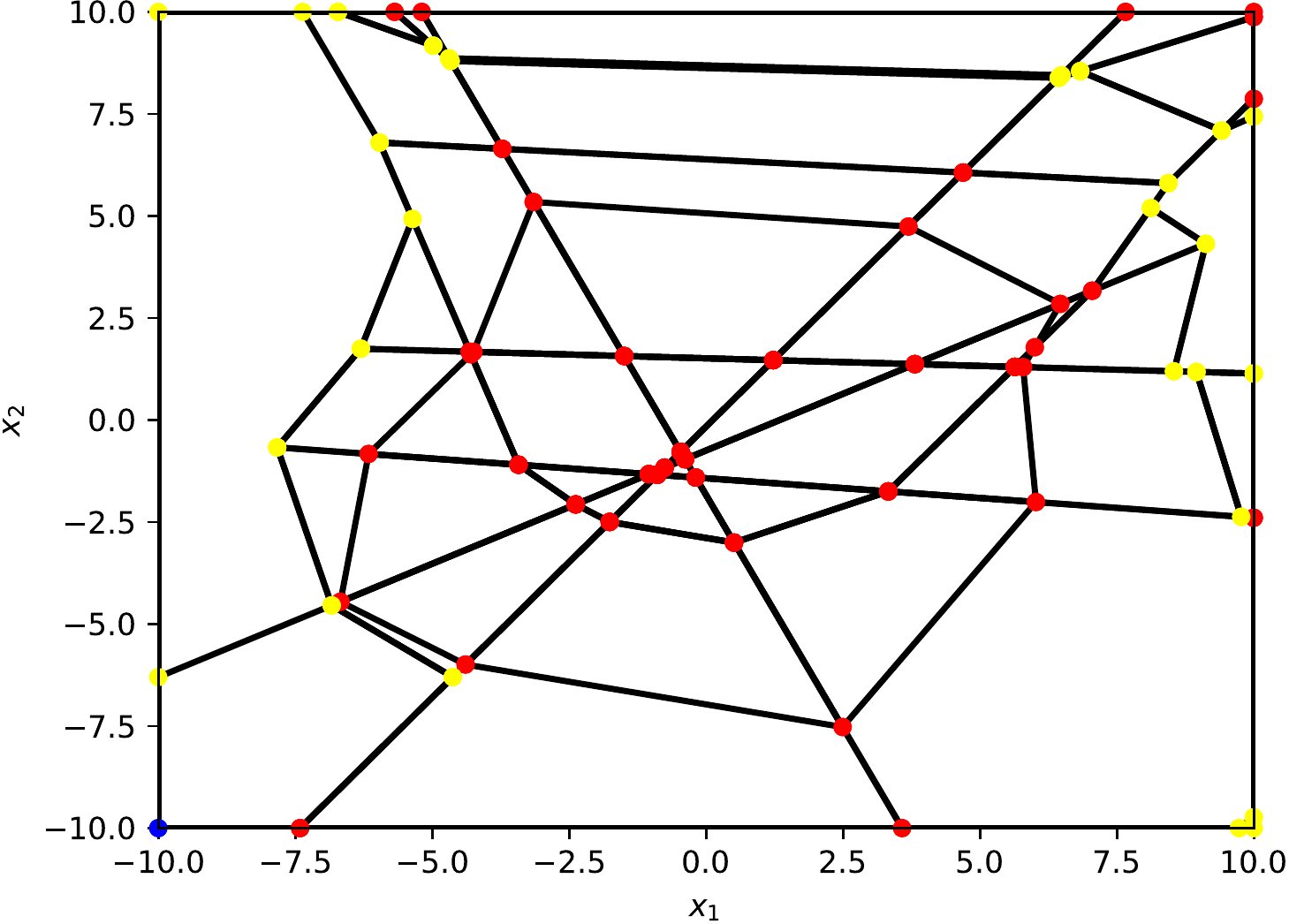}
        \caption{$f^2$}
        \label{fig:skeleton2}
    \end{subfigure}
    \begin{subfigure}{2.6cm}
        \centering
        \includegraphics[width=2.6cm]{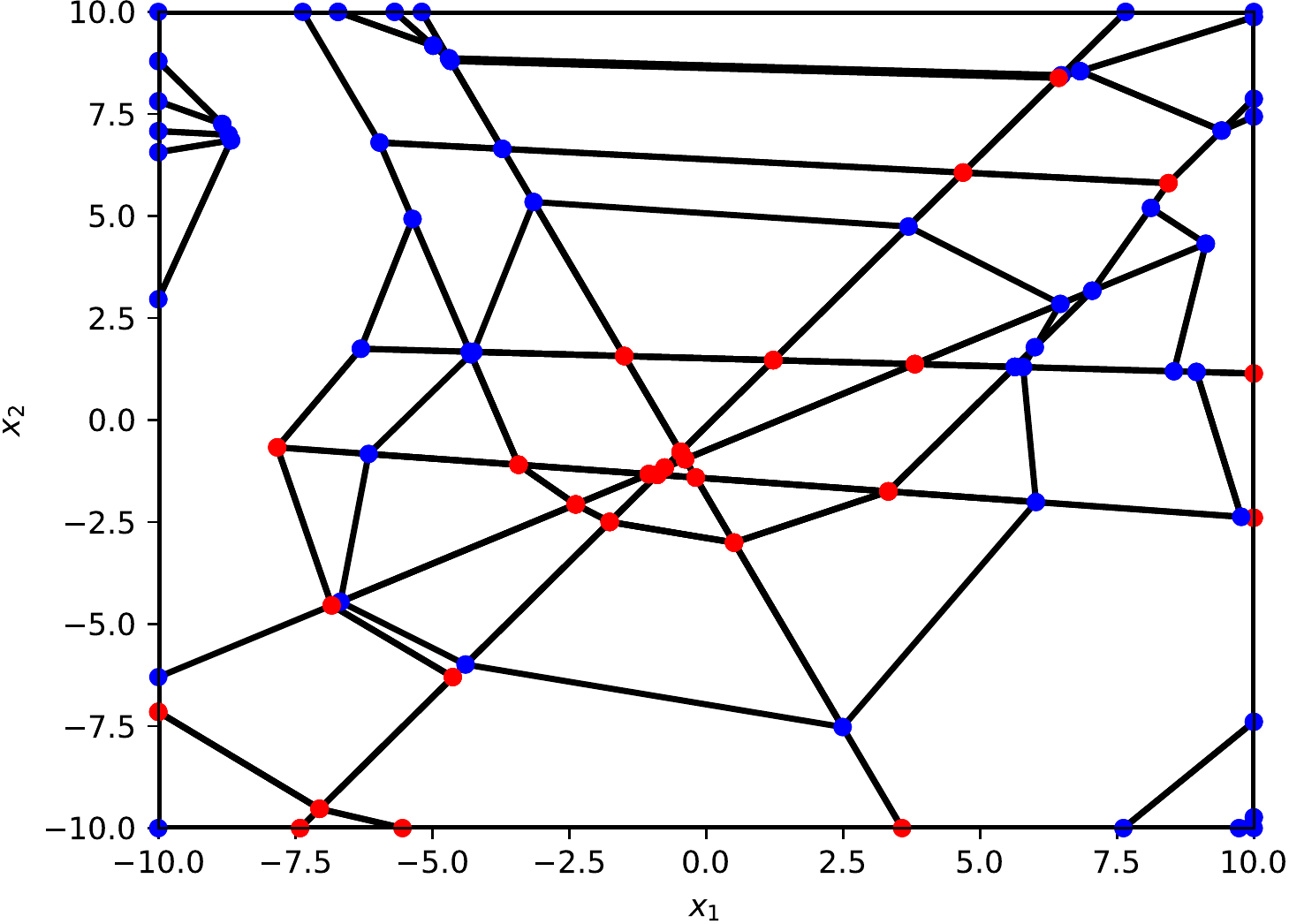}
        \caption{$g = f^1 + f^2 + b$}
        \label{fig:merged_skeleton}
    \end{subfigure}
    \caption{When two functions (a) and (b) are merged together, the result (c) contains vertices and edges of both functions. Here bias $b<0$ is negative, so all yellow vertices turn blue (best viewed on screen).}
    \vspace{-3mm}
    \label{fig:example_of_merging_activations}
\end{figure}

Let's imagine two tessellations $T_1$ and $T_2$ that represent the skeletons of functions $f_i^1$ and $f_i^2$, respectively. Merging those tessellations is equivalent to drawing $T_1$, and then adding all of the lines from $T_2$ that have not already been drawn (Fig. \ref{fig:example_of_merging_activations}). This can be done by going through all linear regions $lr$ in $T_1$, and finding linear regions in $T_2$ that intersect $lr$. Those intersections must have a Lebesgue measure for $n = n_0$, as they will become the tiles forming the tessellation of the merged skeleton. So, for the function in Fig. \ref{fig:example_of_merging_activations}, the intersection must not be a line or a point. Since this tessellation represents higher dimensional functions, we also need to update the value (in the output space) of each vertex ($0$-face) of the skeleton. Hence, given any vertex $v$ in the intersection, we change its value to the sum of values that this vertex has in $f_i^1$ and $f_i^2$ ($v.value = f_i^1(v) + f_i^2(v)$).

\subsection{Additional Insights}
\label{subsec:example_skelex}

Merging activations does not introduce any new critical points. That is why the pre-activations of the given layer produce the same tessellation of the input space. The only exception would be when the weights are equal to $0$, as that reduces the number of activations that are merged. That is why it is highly likely that all $k$ membership functions have the same skeleton, because ReLU is not applied to the output neurons.

We also find that the learned functions tend to produce a significant amount of parallel and almost-parallel line segments visible in Fig. \ref{fig:example_of_applying_relu}, and \ref{fig:BoundEx_examples}. This most likely happens because the activations of the first hidden layer play a pivotal role in forming the skeleton. Therefore, a subspace of the critical points from the activation of a neuron from the first hidden layer appears in many future activations, but each neuron might translate it slightly forming a series of parallel-looking lines. Those lines can be then slightly altered by ReLU. The majority of those lines do not play any role in the creation of the decision boundary.

\begin{figure}[t]
    \begin{subfigure}{4cm}
        \centering
        \includegraphics[width=4cm]{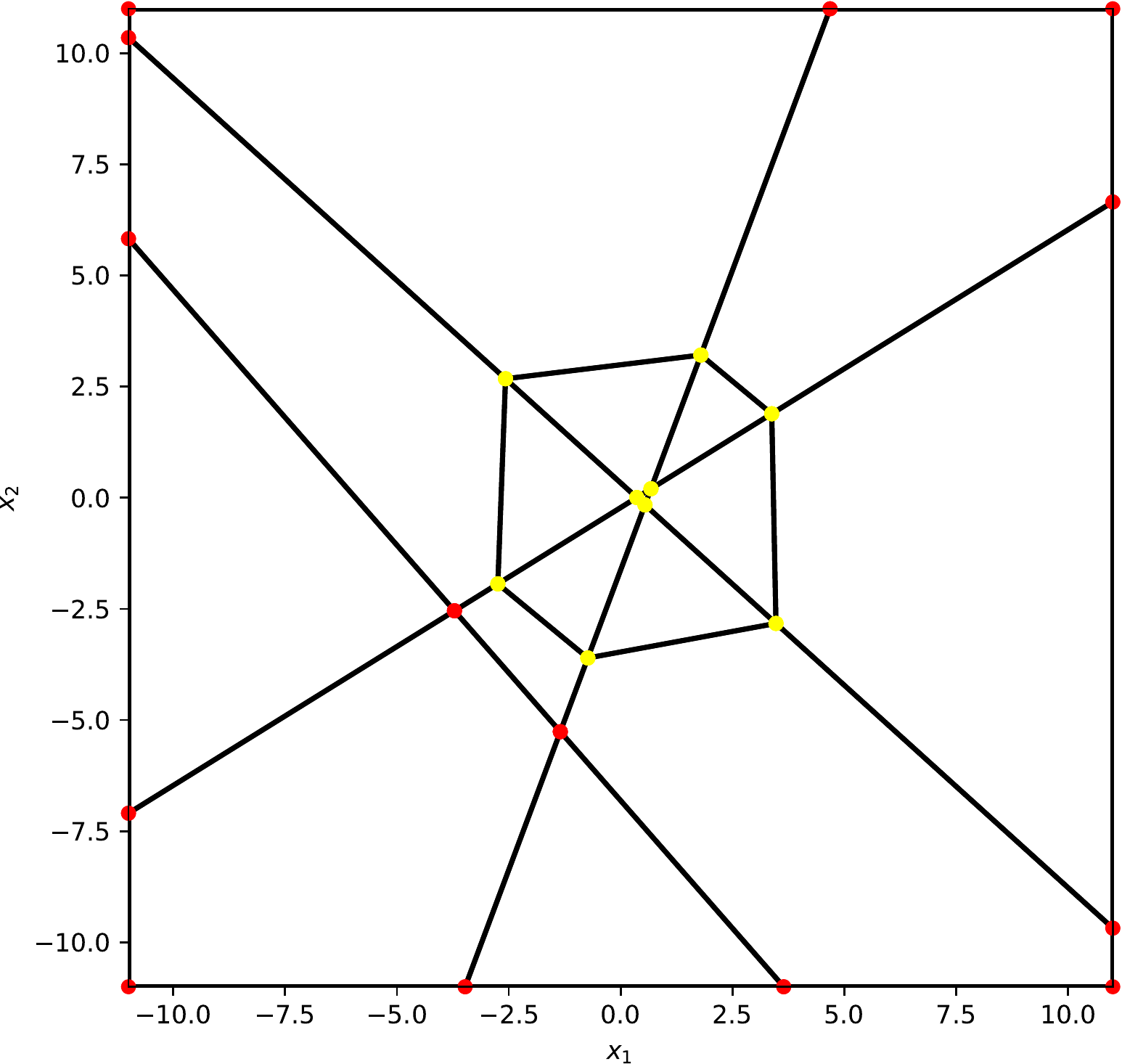}
        \caption{Activation regions}
        \label{fig:activation_regions}
    \end{subfigure}
    \begin{subfigure}{4cm}
        \centering
        \includegraphics[width=4cm]{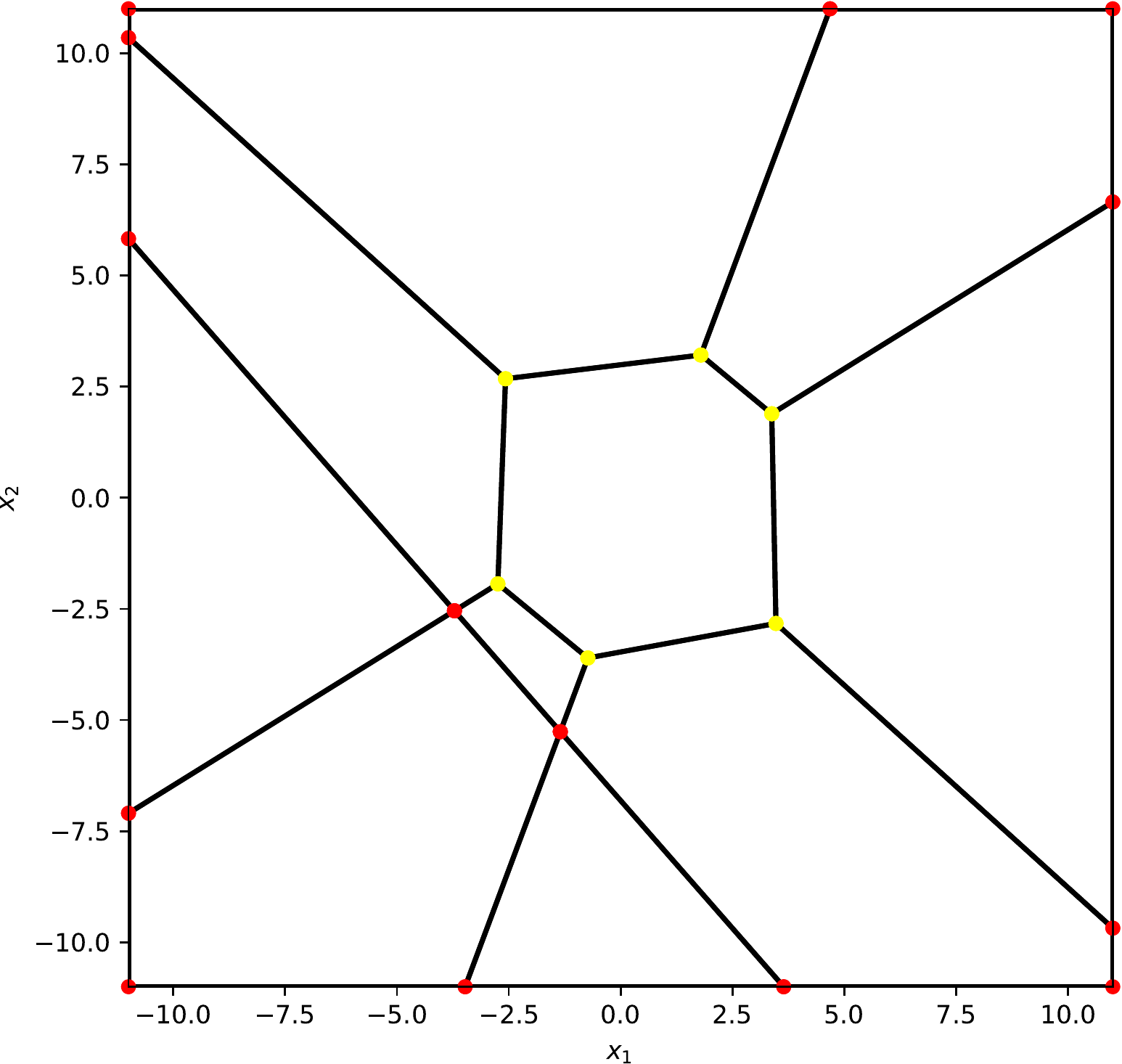}
        \caption{Linear regions}
        \label{fig:linear_regions}
    \end{subfigure}
    \caption{Linear regions and activation regions are not the same. Regions that share gradient can be merged to transform activation regions (a) into linear regions (b).}
    \label{fig:activation_vs_linear_regions}
\end{figure}

Hanin and Rolnick \cite{hanin2019complexity, hanin2019deep} were the first ones to highlight the difference between linear regions and activation regions. In literature, those two terms were often used interchangeably, even though they have different definitions. The activation region is a collection of points in the input space that produce the same activation sequence, whereas the linear region is a subspace of the domain of the membership function where the gradient does not change. The difference between those two notions is clearly visualized in Fig. \ref{fig:activation_vs_linear_regions}. Interestingly, we noticed that the membership functions usually contain the same amount of activation regions as linear regions. The reason for this might be that the activation regions that are removed by ReLU in one neuron can survive in other other neurons. After all, pre-activations of the given layer produce the same tessellation of the input space, so the activation region would have to be removed in all neurons to completely vanish. In other words, working only on activation regions should not change the result, but would make the computational complexity bigger.

\begin{algorithm}
    \caption{MergeActivations}\label{alg:merge_activations}
    \begin{algorithmic}
        \Require $S^l \gets \{S_l^1, ..., S_l^{n_l}\}$  \Comment{$n_l$ skeletons of $l^{th}$ layer}
        \State $current\_S \gets S_l^1$
        \For{$i \in \{2, 3, ..., n_l\}$}
            \State $new\_S \gets$ empty
            \For{$lr_1 \in current\_S$, $lr_2 \in S_l^i$}
                \State $intersection \gets lr_1 \cap lr_2$
                \If{$intersection$ is $n_0$-dimensional}
                    \For{$v \in intersection.vertices$}  
                        \State $v.value \gets lr_1.v.value + lr_2.v.value$
                    \EndFor
                    \State $new\_S.add(intersection)$
                \EndIf
            \EndFor
            \State $current\_S \gets new\_S$
        \EndFor
        \State \Return $new\_S$
    \end{algorithmic}
\end{algorithm}

\section{BoundEx}
\label{sec:4}

Skeletons extracted by SkelEx can be used to analytically calculate the decision boundary. To do so, we need to devise a variation of the $\argmax$ function that uses skeletons of whole functions as an argument rather than a vector of real numbers. We call this variation of $\argmax$ BoundEx (Algorithm \ref{alg:boundex}).

\subsection{BoundEx Algorithm}

Let $f^1, f^2, ..., f^k$ be the membership functions returned by SkelEx, $T$ be the tessellation produced by them, and $lr$ be one of the tiles of $T$. When applying BoundEx to $lr$ we find two cases, and we are only interested in the second case as it indicates the change of membership:
\begin{itemize}
    \item \textbf{Case 1.} $\forall \mathbf{v}\in V \; \forall j\in{1, 2, ..., |V|)} \;\;f^i(\mathbf{v}) >= f^j(\mathbf{v})$, if we let $V$ be the set of vertices of given linear regions
    \item \textbf{Case 2.} $\exists \mathbf{v}\in V \; \exists j_1\in{1, 2, ..., |V|}$ s.t. $f^i(\mathbf{v}) > f^{j_1}(\mathbf{v})$, and $\exists \mathbf{v}\in V \; \exists j_2\in{1, 2, ..., k}$ s.t. $f^i(\mathbf{v}) < f^{j_2}(\mathbf{v})$, with $V$ the same as above
\end{itemize}

\begin{figure*}
    \centering
    \vspace{-5mm}
    \includegraphics[width=3.4cm]{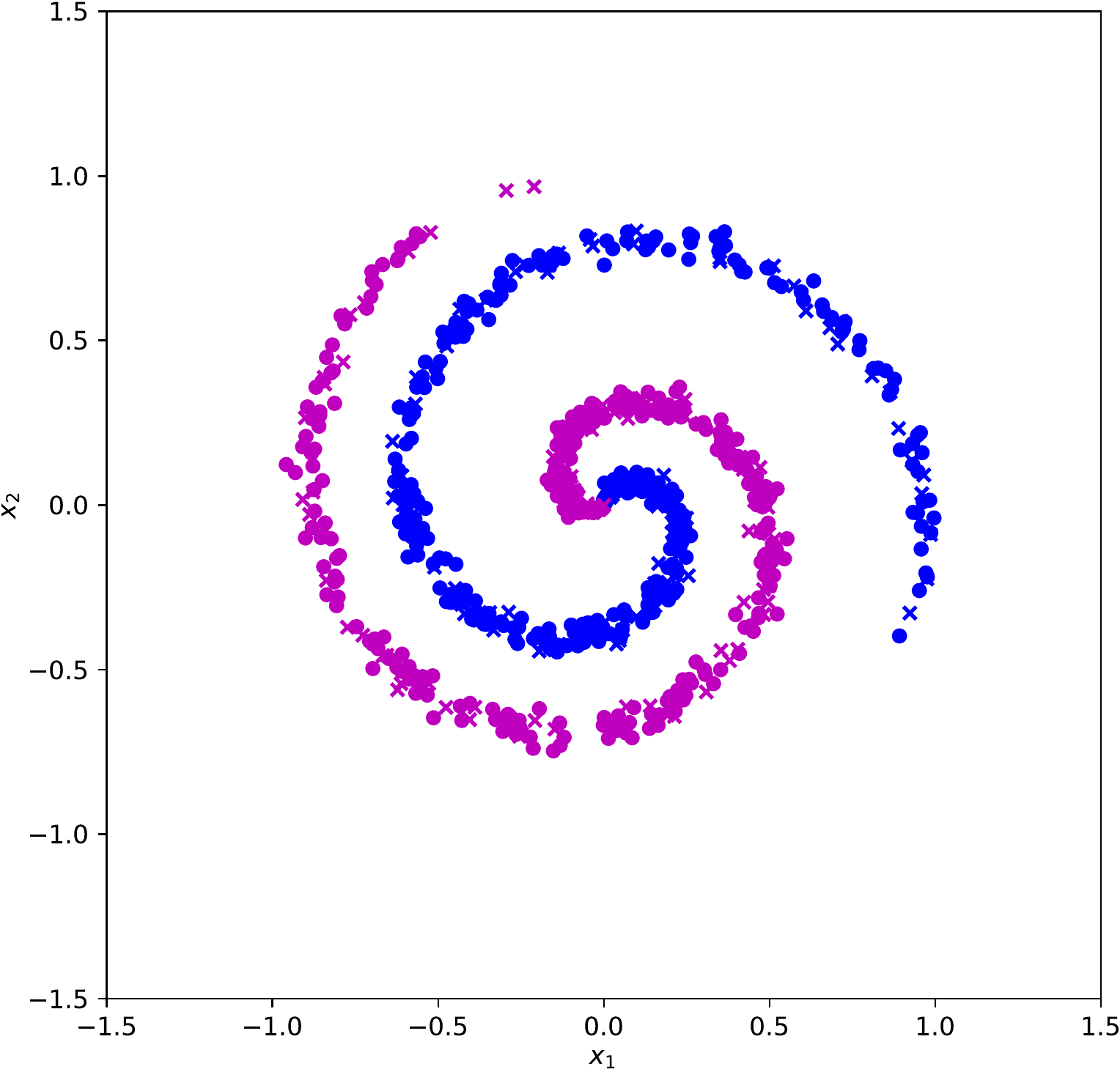}
    \includegraphics[width=3.4cm]{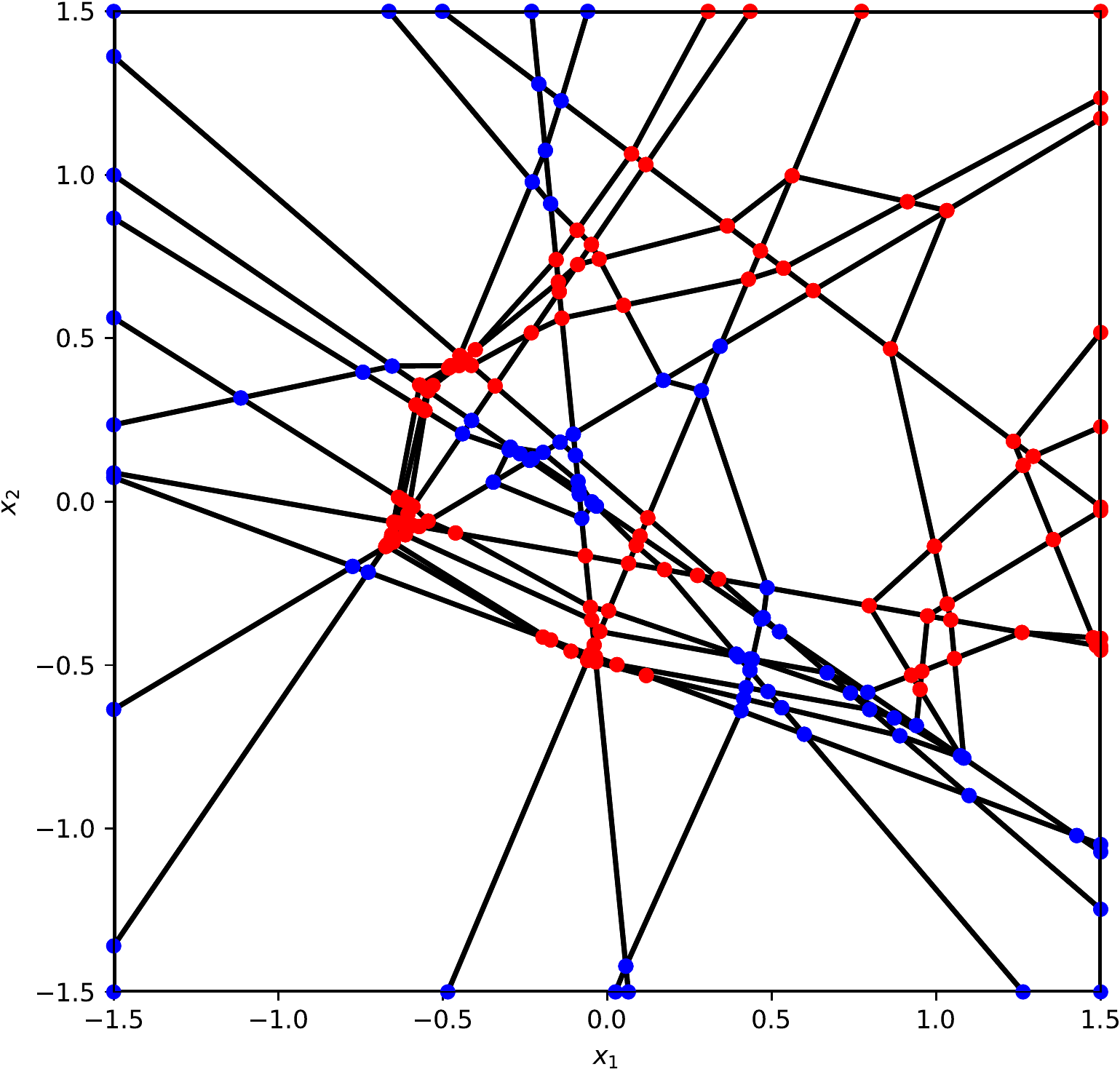}
    \includegraphics[width=3.4cm]{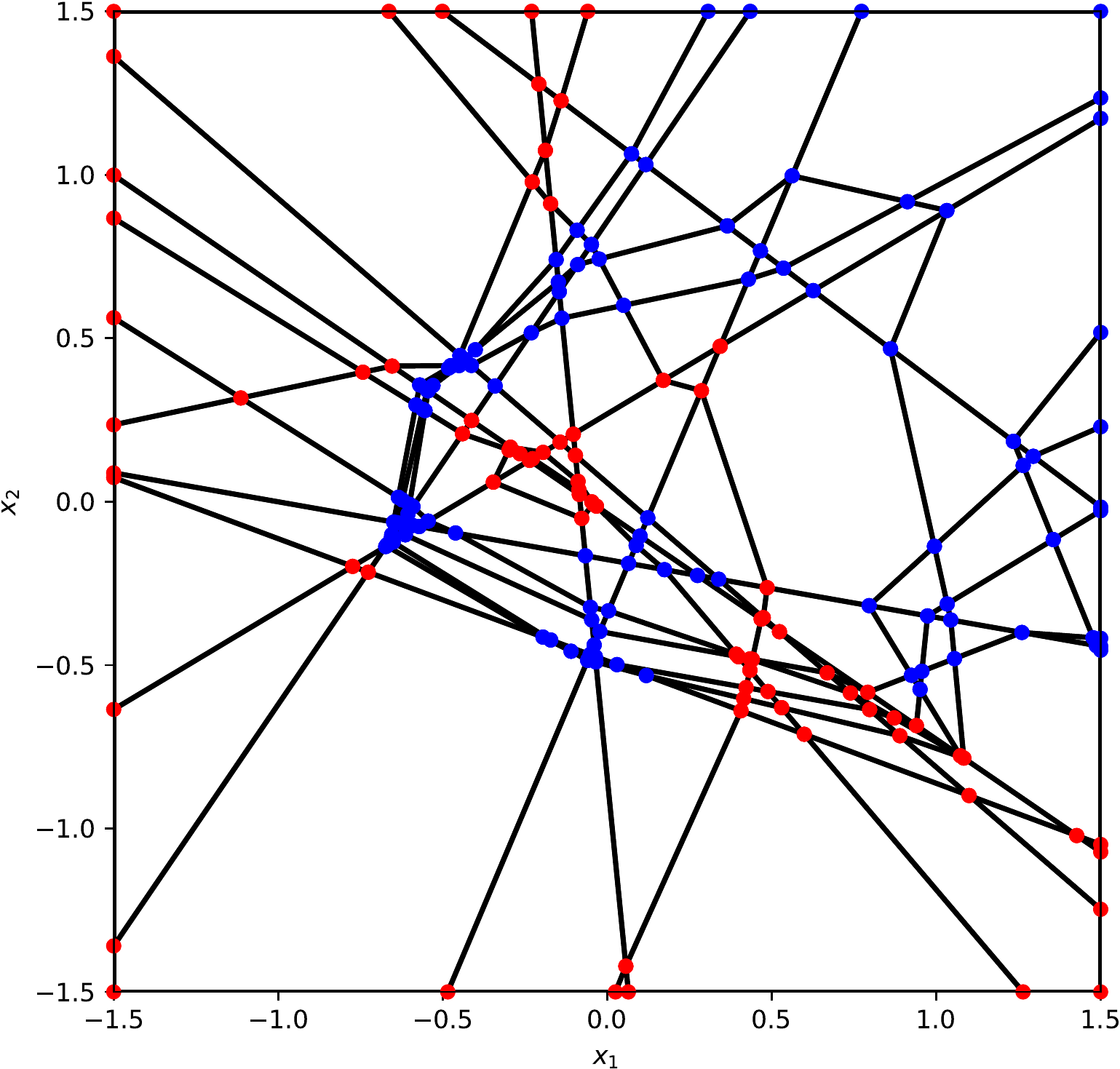}
    \includegraphics[width=3.4cm]{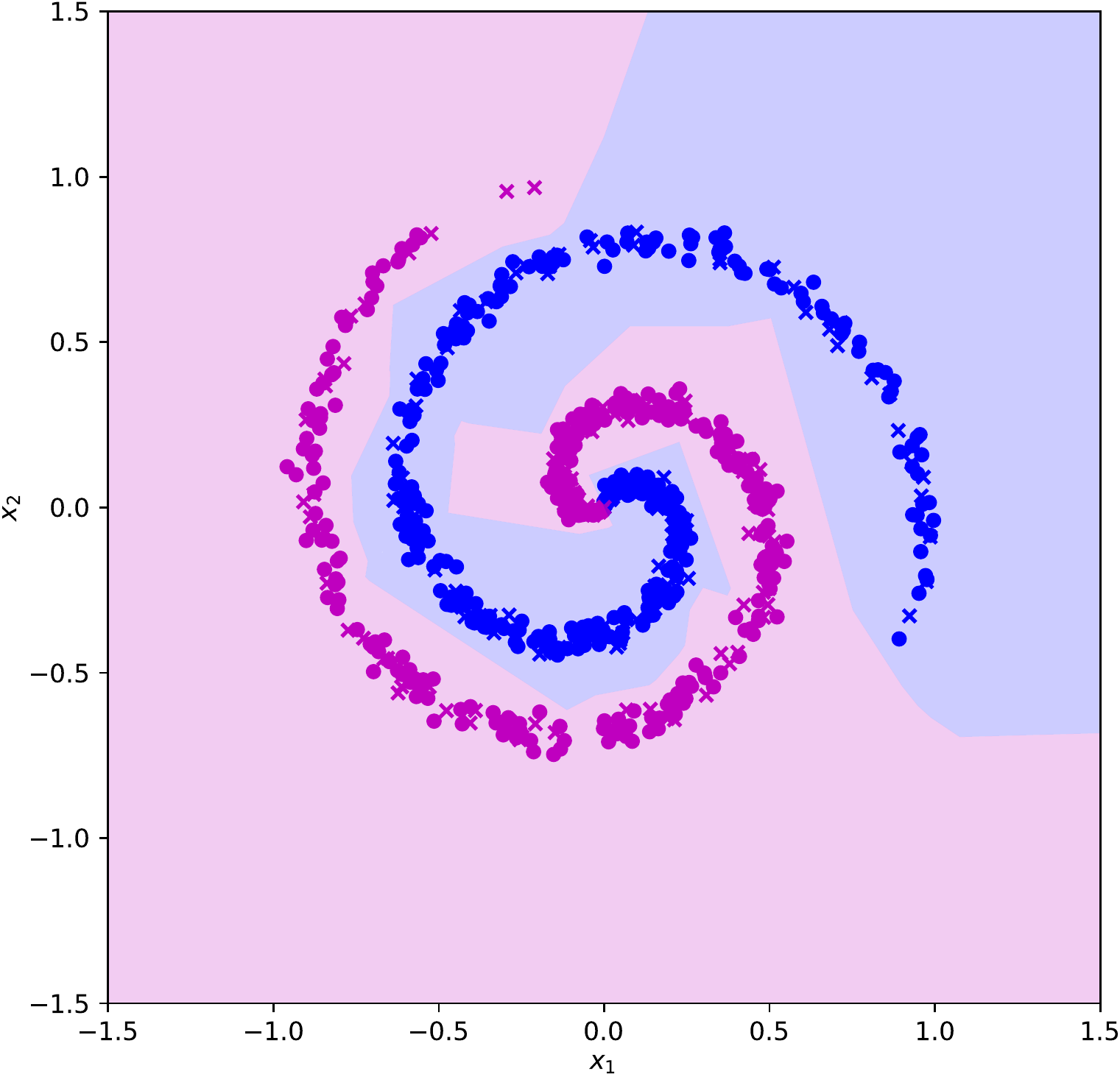}
    \includegraphics[width=3.4cm]{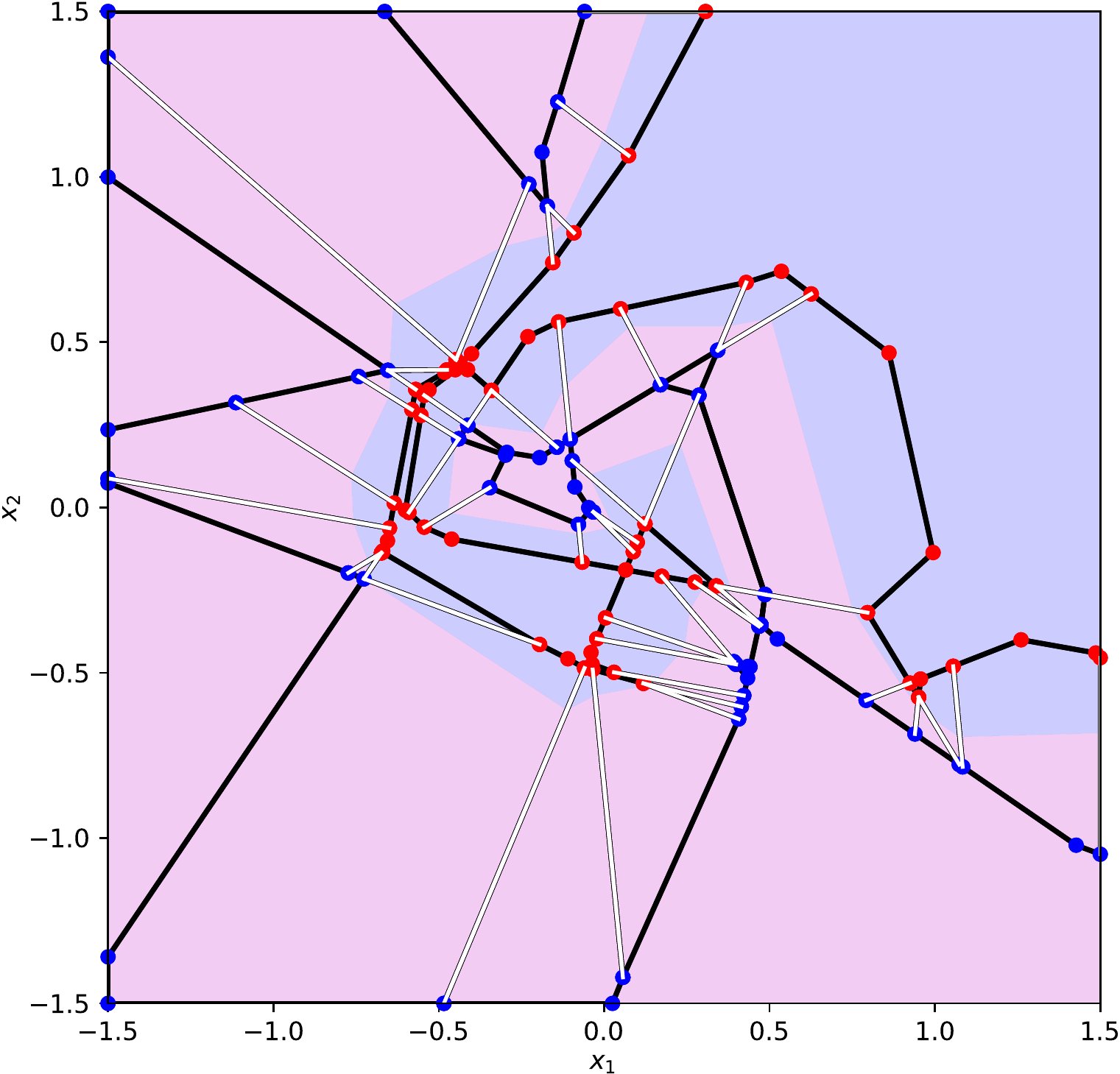}
    \centering
    \includegraphics[width=3.4cm]{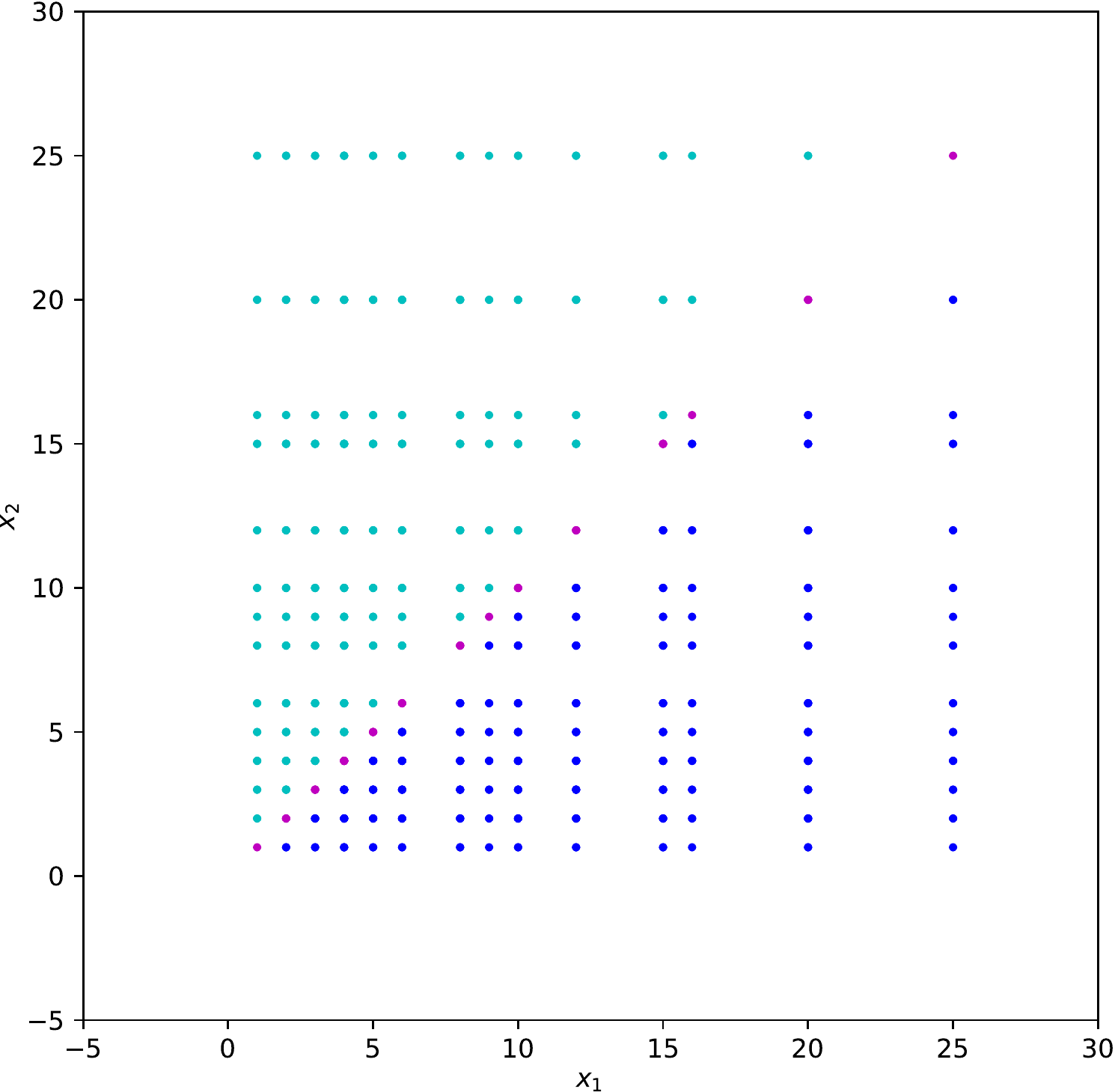}
    \includegraphics[width=3.4cm]{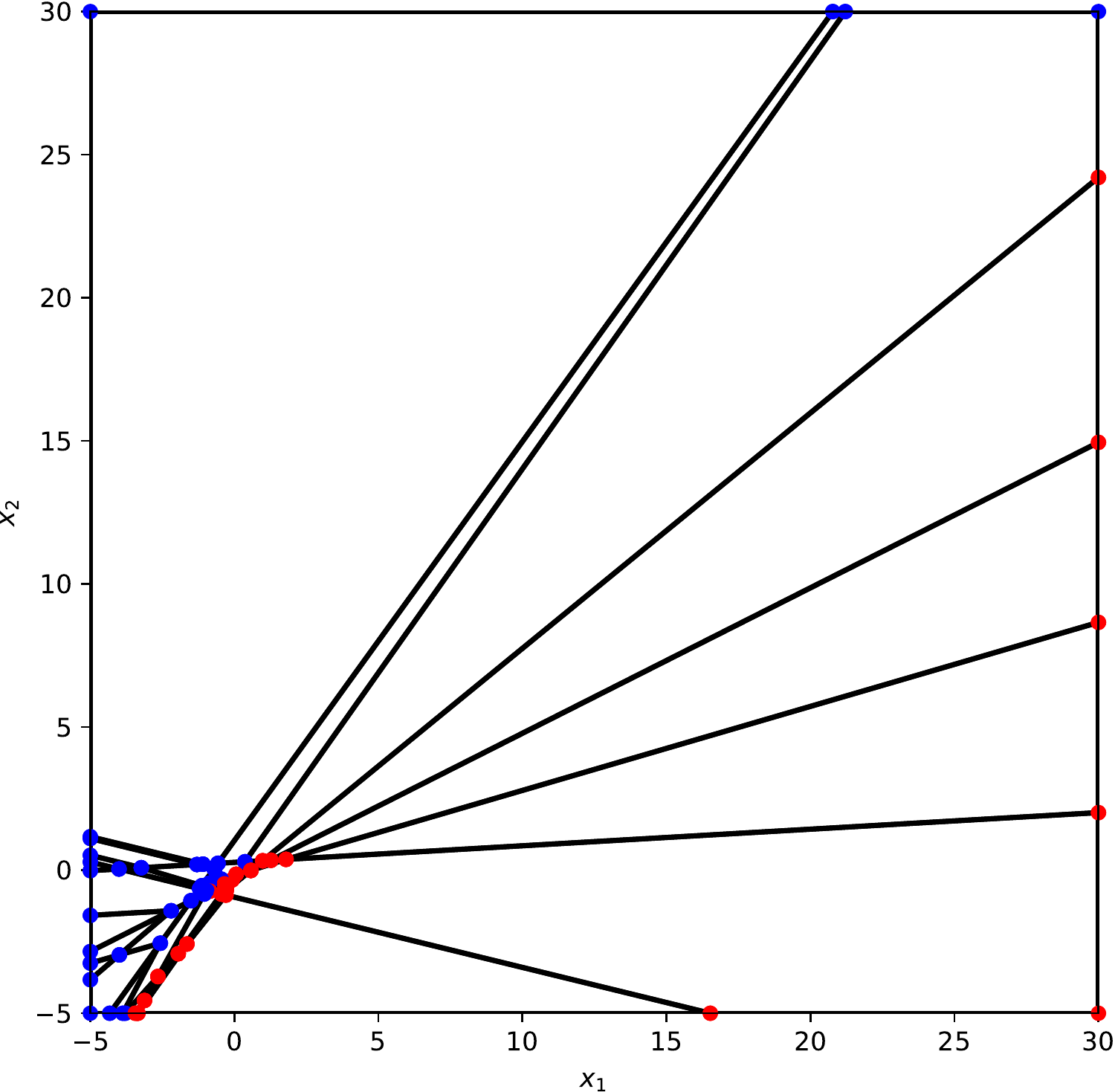}
    \includegraphics[width=3.4cm]{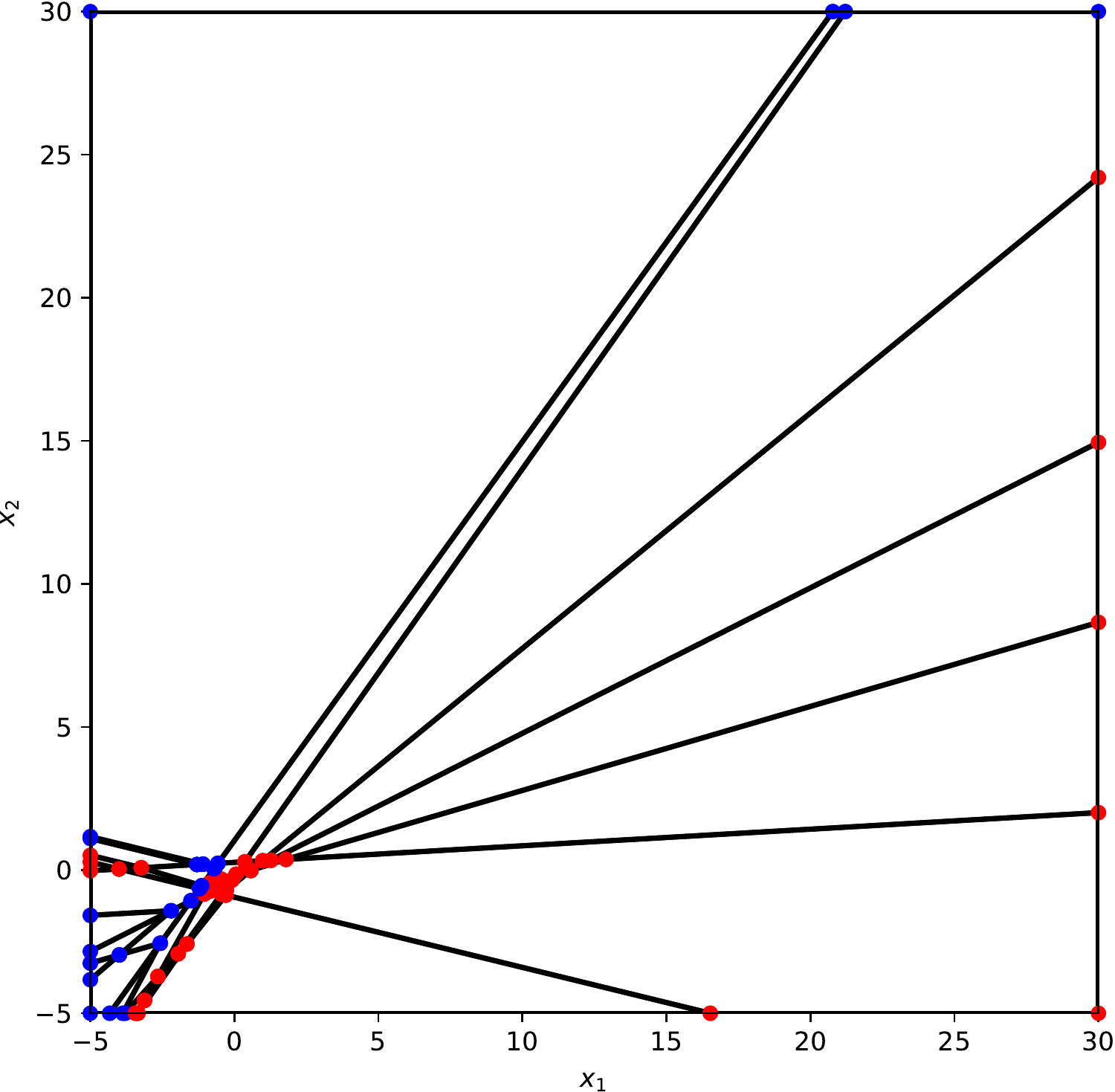}
    \includegraphics[width=3.4cm]{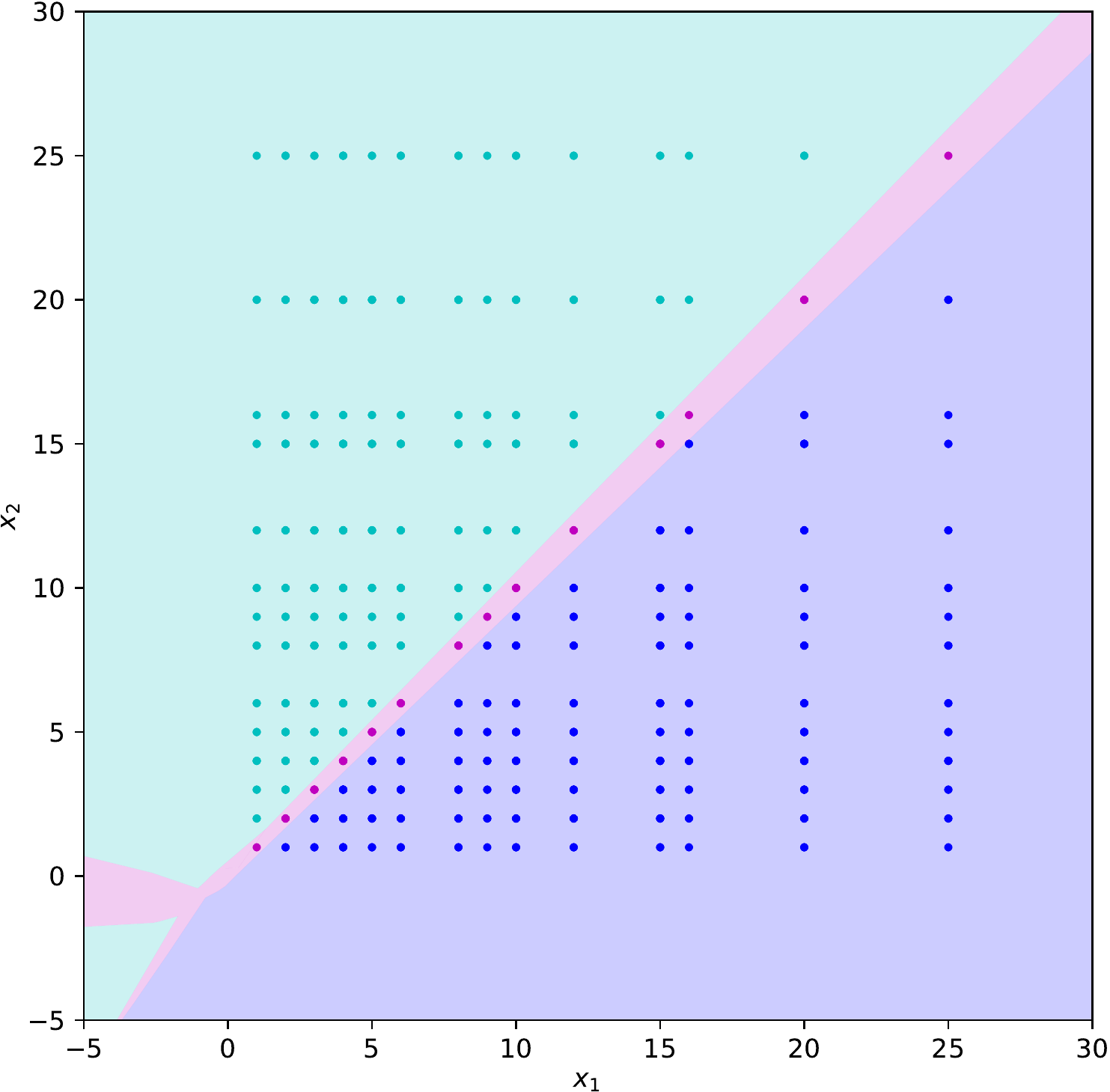}
    \includegraphics[width=3.4cm]{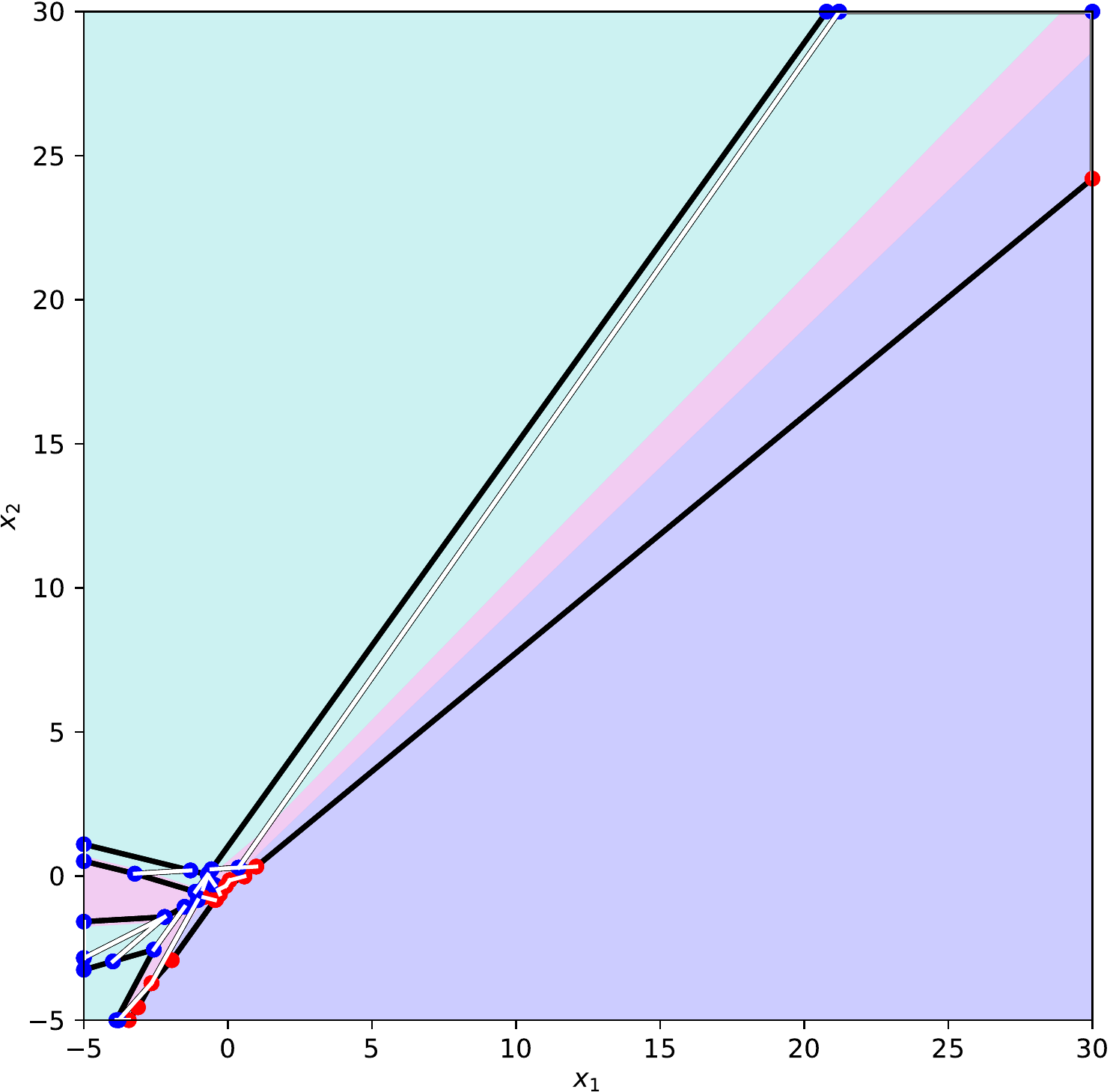}
    \caption{When give weights and biases of ReLU NN trained on data from col. $1$ (row $1$ - toy; row $2$ - Balance Scale), SkelEx extracts the skeletons of learned membership functions (col. $2$ for $f^1$, col. $3$ for $f^2$). BoundEx takes those and extract the decision boundary (col. 4). If we removed the unnecessary linear regions the tessellation would become simpler (col. 5).}
    \vspace{-3mm}
    \label{fig:BoundEx_examples}
\end{figure*}

The decision boundary that passes through $lr$ divides it into up to $k$ membership polygons. To understand what this implies, let $v$ be one of the vertices of $lr$, and $f^i$ be the function such that $f^i = \argmax(f^1(v), f^2(v), ..., f^k(v))$. Now consider any other function $f^{i\_1}$ from the set of membership functions $\mathbf{f}$. If $f^{i\_1}$ intersects $f^i$ within $lr$, then it divides $lr$ into two regions - the one where $f^i$ is larger ($pos\_reg$), and the one where $f^{i\_1}$ is larger ($neg\_reg$). If this was a two-class problem, then $pos\_reg$ would be the polygon enclosing all points that are classified by the model as belonging to class $i$. Now let's take another function $f^{i\_2}$. If $f^{i\_2}$ intersects $f^i$ inside of the $pos\_reg$, then, just as before, $pos\_reg$ gets divided into two regions. We are not interested in the intersection between $f^{i\_2}$ and $f^i$ that happens outside of $pos\_reg$. That is because, for $neg\_reg$, $f^{i\_1}$ is larger than $f^i$. So to determine the membership of points within $neg\_reg$, we must compare $f^{i\_2}$ with $f^{i\_1}$, not with $f^i$. If we continue this process until we go through all $k$ functions we will obtain all \textit{membership polygons}. Those polygons define the set of points that are classified as belonging to the same class by the trained neural network. We do the same for all other tiles from $T$, and then merge the neighboring membership polygons of the same class. The decision boundary can be found at the interception of those polygons.

\subsection{Examples and Limitations}
\label{subsec:1}

To present the capabilities of BoundEx, we train it on a toy datasets, and the Balance Scale dataset from UCI \cite{Dua:2019} (Fig. \ref{fig:BoundEx_examples} col. 1). For the Balance Scale dataset, we reduce the dimensionality of the data to 2 by introducing variables $x_1 = weight_l \times distance_l$, and $x_2 = weight_r \times distance_r$. We train a ReLU NN on each of those datasets to $\sim100\%$ accuracy. We then pass the learned weights and biases to SkelEx to produce $k$ skeletons of membership functions. As mentioned in Sec. \ref{subsec:example_skelex}, we see that both membership functions produce the same tessellation of the input space. Those membership functions generate the decision boundaries, and the decision polygons (col. $4$). Now, when given a data sample to classify, rather than performing forward propagation, we can just check in which polygon it lies. Hence, BoundEx transforms the classification problem into a Point-in-Polygon problem.

\begin{algorithm}
    \caption{BoundEx}\label{alg:boundex}
    \begin{algorithmic}
        \State $mp \gets$ hash map  \Comment{\textit{membership polygons}}
        \For{$lr \in T$}
            \State $v \gets lr.vertices[1]$
            \State $base_f = f^{\text{argmax}([f^1(v), f^2(v), ..., f^k(v)])}$
            \State $cmp[$index$(base_f)] = lr$  \Comment{current mp}
            \For{$f^i \in \mathbf{f}$ s.t. $i \notin cmp.keys$}
                \For{$key \in cmp.keys$}
                    \For{$reg \in cmp[key]$}
                        \State $pos\_reg, neg\_reg = reg.split(f^i)$ 
                        \State remove $neg\_reg$ from $cmp[key]$
                        \State $cmp[i]$.append$(neg\_reg)$
                    \EndFor
                \EndFor
            \EndFor
            \For{$key \in cmp.keys$}
                \State $mp[key].extend(cmp[key])$
            \EndFor
        \EndFor
        \State merge neighboring membership polygons of the same class
        \State \Return $dp$
    \end{algorithmic}
\end{algorithm}

It is common knowledge that NNs are overparameterized, and because of that learn a lot of unnecessary knowledge. This is clearly visible in columns $2$ and $3$ of Fig. \ref{fig:BoundEx_examples}. The skeleton contains multum of faces that are have no impact on the production of the decision boundary. If we remove all the unnecessary linear regions (and adjust the values of the remaining ones accordingly), the skeleton would become significantly simpler. In fact, we should be able to separate the faces of the skeleton that pass through the vertices of the decision boundary (white lines on col $5$, of Fig. \ref{fig:BoundEx_examples}), and use them to craft a smallest possible skeleton that generates given decision boundary.

The activations of the first hidden layer are composed of two $n_0$-dimensional linear regions\footnote{The only exception are the cases where the hyperrectangle is too small, and so there is only one linear region.}. Since an $n_0$-dimensional linear region requires $2^n$ vertices to encode, we need $n_1\times2^{n_0}$ vertices to encode all of the skeletons from the first hidden layer. This number increases as SkelEx progresses through the hidden layers, making the algorithm expensive for high-dimensional data. Experiments in this work contain were performed only on 2D data, because in our implementation we use GEOS library \cite{GEOS} that does not support polygon operations for higher dimensions. 

\section{Conclusion}
\label{sec:6}

In this work, we introduced SkelEx and BoundEx. SkelEx is an algorithm that, given weights and biases of pre-trained ReLU NN, extracts the skeleton of the learned membership functions. It operates in two steps, where it iteratively applies ReLU to the pre-activations to calculate activations and then merges the activations to calculate the pre-activations of the neurons of the next layer. We have shown that SkelEx provides a very natural visualization method, which yields nice results. We have also introduce BoundEx, the first analytical method that extracts the decision boundary learned by ReLU NNs.

%
\bibliographystyle{IEEEbib}
\bibliography{paper}

\end{document}